\documentclass{trbunofficial}
\usepackage{graphicx}
\usepackage{subcaption}
\usepackage{array}
\usepackage{tcolorbox}
\tcbuselibrary{skins,breakable}

\usepackage{wrapfig}
\newcolumntype{C}[1]{>{\centering\arraybackslash}m{#1}}  
\usepackage[hidelinks]{hyperref}
\usepackage{multirow}

\newcount\Comments  
\usepackage{color}
\definecolor{darkgreen}{rgb}{0,0.5,0}
\definecolor{purple}{rgb}{1,0,1}
\newcommand{\kibitz}[2]{\ifnum\Comments=0\textcolor{#1}{#2}\fi}

\AuthorHeaders{Parsa, Li, Kockelman, and Choi}
\title{Video-based Vehicle Surveillance in the Wild: License Plate, Make, and Model Recognition with Self Reflective Vision-Language Models}

\author{%
  \textbf{Pouya Parsa}\\
  Department of Civil, Environmental, and Geo- Engineering\\
  University of Minnesota\\
  parsa025@umn.edu\\
  \hfill\break
  \textbf{Keya Li}\\
  Department of Civil, Architectural and Environmental Engineering\\
  The University of Texas at Austin \\
  keya\_li@utexas.edu\\
  \hfill\break
  \textbf{Kara M. Kockelman, Ph.D., P.E.}\\
  Department of Civil, Architectural and Environmental Engineering\\
  The University of Texas at Austin \\
  kkockelm@mail.utexas.edu\\
  \hfill\break
  \textbf{Seongjin Choi, Ph.D., Corresponding Author}\\
  Department of Civil, Environmental, and Geo- Engineering\\
  University of Minnesota\\
  chois@umn.edu\\
  \hfill\break
}

\TotalWords{5390}

\begin{document}

\maketitle

\section{Abstract}

Automatic license plate recognition (ALPR) and vehicle make and model recognition underpin intelligent transportation systems, supporting law enforcement, toll collection, and post-incident investigation. Applying these methods to videos captured by handheld smartphones or non-static vehicle-mounted cameras presents unique challenges compared to fixed installations, including frequent camera motion, varying viewpoints, occlusions, and unknown road geometry. Traditional ALPR solutions, dependent on specialized hardware and handcrafted OCR pipelines, often degrade under these conditions. Recent advances in large vision–language models (VLMs) enable direct recognition of textual and semantic attributes from arbitrary imagery.
This study evaluates the potential of VLMs for ALPR and makes and models recognition using monocular videos captured with handheld smartphones and non-static mounted cameras. The proposed license plate recognition pipeline filters to sharp frames, then sends a multimodal prompt to a VLM using several prompt strategies.
Make and model recognition pipeline runs the same VLM with a revised prompt and an optional self-reflection module. In the self-reflection module, the model contrasts the query image with a reference from a 134-class dataset, correcting mismatches. Experiments on a smartphone dataset collected on the campus of the University of Texas at Austin, achieve top-1 accuracies of 91.67\% for ALPR and 66.67\% for make and model recognition. On the public UFPR-ALPR dataset, the approach attains 83.05\% and 61.07\%, respectively. The self-reflection module further improves results by 5.72\% on average for make and model recognition. These findings demonstrate that VLMs provide a cost-effective solution for scalable, in-motion traffic video analysis.

\hfill\break%
\noindent\textit{Keywords}: Automatic license plate recognition, vision–language models, CLIP image quality assessment, prompt engineering, traffic law enforcement, automated enforcement, self-reflection

\clearpage
\section{Introduction}\label{sec:intro}
Nearly 1.2 million people die each year globally in road traffic crashes~\cite{who2023road}, with over 40,000 fatalities occurring in the United States alone in 2023~\cite{nhtsa2024traffic}. Speeding, reckless driving, and hit-and-run collisions are major contributors to crashes and fatalities, yet the lack of real-time enforcement tools makes preventing them especially difficult. Stationary cameras, while effective, are very expensive to install and maintain, averaging over \$120,000 per location~\cite{ibo2016speedcameras}. Many nations, states, and cities are reluctant to adopt them due to their limited coverage from fixed placements and concerns over privacy. Moreover, drivers often slow down near enforcement points (and speed up downstream, once out of camera view), a phenomenon known as the ``kangaroo effect''~\cite{SafaviNaini2024}. A scalable alternative lies in enabling citizens to assist enforcement by capturing video footage of infractions using smartphones or dashcams. However, leveraging such crowd-sourced data requires robust tools for analyzing low-quality and personal videos, which is a significant technical challenge.

Several nations have developed public reporting platforms to facilitate citizen-driven traffic enforcement. For example, in South Korea, citizens can use the official ``Safety e-Report'' service\footnote{\url{https://www.safetyreport.go.kr/eng/}} to report traffic violations by uploading video evidence directly to law enforcement. In return, some users may receive monetary rewards or recognition, effectively turning smartphones into distributed enforcement tools and encouraging community participation in road safety~\cite{koreaApp2023,choi2023analysis}. In New York City, citizens have been helping enforce diesel-truck idling laws; those submitting 3 minutes of video receive 25 percent of any fine obtained from heavy-truck owners, which is close to \$87.50~\cite{wilson2022idlingtrucks}. Such low-cost programs cost-effectively extend enforcement reach while building social accountability among drivers.

Despite their success, these systems still require users to manually enter important vehicle information, such as the license plate number, make, and model. In many cases, this information is added after the video upload is complete, or left out entirely when the plate is difficult to read. This extra step increases the burden on users and reduces the number of reports that can be successfully verified. Automating the extraction of this kind of information directly from the video would improve ease of use, minimize input errors, and increase the likelihood of valid reports for enforceable actions.

To enable automatic traffic law enforcement from citizen-recorded videos, two core recognition tasks are essential: \textbf{Automated license plate recognition} (ALPR) and \textbf{vehicle make (manufacturer) and model recognition}. These tasks serve as the foundation for identifying and tracking offending vehicles across time and locations. ALPR allows for the extraction of unique vehicle identifiers, enabling citation issuance, database cross-referencing, and ownership tracing. Recognizing the make and model of a vehicle provides an additional layer of verification or confirmation, and is especially valuable in cases of partial plate visibility, occlusion, or modified plates. Together, these two tasks form the minimum viable input required for automating enforcement actions from unstructured video evidence.

ALPR, also known as Automatic Number Plate Recognition (ANPR), is a basic intelligent transportation system (ITS) technology for automated enforcement, enabling the identification of vehicles involved in traffic violations, toll evasion, and criminal activities. It can play a critical role in issuing citations, locating stolen vehicles, and monitoring access points (to paid parking lots, tollways, and high-security events). However, traditional ALPR systems depend on high-resolution cameras and controlled environments, making them unreliable when applied to noisy footage from standard smartphones or legacy closed-circuit television (CCTV) camera systems. In such cases, license plates may occupy only a few pixels or appear blurred or occluded, reducing the effectiveness of conventional optical character recognition (OCR) systems. An illustrative comparison is provided in Figure~\ref{fig:plates}, contrasting an ideal high-resolution US license plate with a low-resolution, blurry plate often seen in real-world video footage.

\begin{figure}[t]
    \centering
    \includegraphics[width=\textwidth]{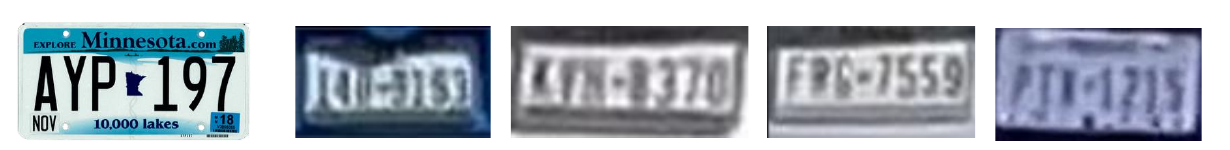}
    \caption{\textbf{Comparison of ideal and real-world low-quality license plates}}
    \label{fig:plates}
\end{figure}

Beyond plate recognition, extracting vehicle make and model further enhances enforcement capabilities by supporting cross-verification, vehicle re-identification, and suspect profiling. But this task typically relies on separate classifiers that are sensitive to viewpoint variations and trained on limited datasets, making them brittle and hard to scale.

Conventional approaches \cite{conv_cnn_ocr, conv_cnn_ocr_1, conv_cnn_ocr_2}, for ALPR typically rely on dedicated OCR engines applied to tightly cropped plate regions, while make and model recognition is handled by specialized CNN-based classifiers trained on curated vehicle datasets \cite{Satar_2018_mmr}, \cite{Li2025smartphone}. These pipelines often assume clean, high-resolution imagery and controlled viewing conditions, which rarely hold in citizen-sourced or dashcam videos. In such scenarios, low resolution, motion blur, obstructions, and unusual angles significantly degrade the performance of traditional classification models. Moreover, maintaining and deploying multiple task-specific models increases system complexity and makes real-time, on-device processing difficult. A practical system therefore needs to be task‑agnostic. In other words, it should take an entire video sequence as input and, without relying on separate specialist modules, return the plate text together with make and model.

Vision–Language Models (VLMs) emerge as a potential solution to this challenge, as they are trained on large collections of image–caption pairs and can therefore leverage a single network to jointly handle visual and textual information.
%
VLMs can recognize objects and comprehend natural language without switching between different modules thanks to the vision encoder's conversion of pixels inside the model into internal features that the text encoder can read instantly. 
This integration allows the model to combine visual understanding and linguistic reasoning seamlessly.
For example, in the context of license plate, make, and model recognition, visual understanding can be used to spot a vehicle and read a partly blurred plate, while reasoning can be used to combine those observations to respond to a prompt, such as ``What make and model is the car, and what does the plate say?'' Because prompts steer the model at run‑time, the same network can tackle many tasks with little or no extra training.

In this study, we develop a license plate recognition pipeline in which high-quality frames are first selected using a frame-quality ranking procedure. A multimodal (image + text) prompt is then constructed and passed to a VLM. We employ different prompting strategies for ablation studies. We evaluate the proposed pipeline using four recent VLMs: GPT-4o \cite{openai2024gpt4o}, Llama 3.2-Vision \cite{meta2024llama32}, LLaVA \cite{liu2023visual}, and MiniCPM-V \cite{yao2024minicpmv}.

For vehicle make and model recognition, we adopt a similar pipeline, with an additional self-reflection module designed to improve make/model recognition robustness. 
Self-reflection module prompts the VLM to verify and, when necessary, revise its own output based on a retrieved image. The module functions in three steps. First, it retrieves the most visually similar vehicle image from a web-crawled dataset of different make/models. Next, it instructs the VLM to compare this retrieved reference image with its initial prediction (and original image). Finally, it returns an updated answer only when the visual evidence suggests a change. 
In this setting, we focus on GPT-4o and Llama 3.2-Vision. 
Experiments show that self-reflection yields a modest but consistent accuracy gain across all tested VLMs and both evaluation datasets.
%



\section{Related Works}
Traditional ALPR systems typically employ a multistage pipeline that involves license plate detection, character segmentation, and character recognition, often relying on handcrafted features and OCR techniques \cite{s21093028}. 
While these techniques work well in controlled settings with high-resolution cameras and ideal lighting, they often perform worse in the presence of issues like motion blur, occlusions, different viewpoints, and poor image quality, which are common in video recorded by portable devices like smartphones.

Recent advances in deep learning have significantly improved ALPR robustness. For example, LPRNet \cite{zherzdev2018lprnetlicenseplaterecognition} introduces a lightweight convolutional neural network (CNN) that performs end-to-end license plate recognition, achieving real-time performance with high accuracy on standard datasets. Despite these advances, deep learning-based ALPR systems still struggle with the variability and noise inherent in unconstrained environments, such as those encountered in smartphone-captured video.

\subsection{Vehicle Make and Model Recognition}
Parallel to ALPR, vehicle make and model recognition has gained traction in applications like traffic monitoring, surveillance, and autonomous driving. Deep learning approaches have been proven effective in this domain. For instance, Deep Learning Vehicle Classification \cite{Satar_2018} employs CNNs to classify vehicles based on visual features, demonstrating high accuracy on fine-grained vehicle datasets. These systems, however, typically require large annotated datasets and may not generalize well to diverse real-world conditions, such as varying lighting or partial occlusions.

\subsection{Vision-Language Models (VLMs)}
The emergence of large vision–language models (VLMs) has introduced a paradigm shift in computer vision by enabling joint reasoning over visual and textual information. Representative examples include CLIP~\cite{radford2021learningtransferablevisualmodels}, which learns aligned image–text representations to support a wide range of vision tasks. CLIP, GPT-4o \cite{openai2024gpt4o}, and LLaVA \cite{liu2023visual} are pre-trained on extensive datasets of image-text pairs, enabling them to perform a wide range of tasks, including zero-shot classification and text recognition without task-specific training. These models excel in understanding complex scenes and can generate textual descriptions from visual inputs, making them well-suited for tasks like ALPR and vehicle recognition.

\citet{nagaonkar2025benchmarkingvisionlanguagemodelsoptical} recently examined the use of VLMs for OCR tasks in dynamic environments. Their study benchmarked models including Claude-3 \cite{anthropic_claude3_2024}, Gemini-1.5 \cite{gemini15_2024}, and GPT-4o \cite{openai2024gpt4o} on video frames sourced from various domains such as code editors and advertisements. The results indicate that VLMs can outperform conventional OCR systems, such as EasyOCR and RapidOCR, in complex scenarios. However, the study highlights ongoing limitations, including occasional hallucinated outputs and sensitivity to stylized text.

Specifically for ALPR, AlDahoul et al.'s Advancing Vehicle Plate Recognition manuscript \cite{aldahoul2024advancingvehicleplaterecognition} demonstrates the application of VLMs to recognize license plates under challenging conditions, such as low illumination, motion blur, and tightly packed characters. The authors fine-tuned a VLM (PaliGemma) to develop VehiclePaliGemma, achieving 87.6\% accuracy on a Malaysian license plate dataset consisting of 258 labeled images. Their method leverages multitask prompting to identify plates in complex scenes involving multiple vehicles with different colors and models. While effective, their reliance on model fine-tuning introduces an additional barrier for adoption, since it requires access to training infrastructure and expertise. In contrast, our approach employs zero-shot prompting with off-the-shelf VLMs, avoiding the need for retraining and making it easier to adapt and deploy in diverse ALPR settings.

\subsection{Self-reflection and Iterative Refinement in VLMs}
To further enhance reliability, recent work has explored \emph{self-reflection strategies} in VLMs, allowing the model to verify and correct its own predictions in multi-step processes. For example, a retrieval-augmented test-time adaptation (RA-TTA) method retrieves reference images from a large external database for each query image and uses them to refine the model’s initial prediction~\cite{ratta2024openreview}. This provides an external visual check on the output by comparing the query with similar images, much like our approach of cross-verifying the VLM’s guess with a database image of the predicted vehicle. Other approaches focus on internal self-critique, the R3V framework prompts a multimodal model to reflect on its reasoning chain and refine any flawed rationale by comparing several reasoning candidates~\cite{r3v2024arxiv}, which helps the model deliver a more accurate answer. In the domain of large language models, the \emph{Self-Refine} technique has similarly shown that a model can iteratively improve its outputs by generating an initial response, then providing feedback on that response, and refining it, all without additional training data or supervision~\cite{selfrefine2023arxiv}. Together, these efforts underscore a broader trend in introducing a feedback loop, whether through external retrieval or internal reflection, which enables vision-language systems to validate and refine their predictions and leads to more robust and accurate recognition in complex scenarios.

\section{Methodology}

We present the details of the proposed license plate recognition pipeline and the make and model recognition pipeline in this section. They are based on similar ideas; however, they have distinct structures due to the nature of the task to be dealt with. The license plate recognition pipeline has two submodules: Input Processing and VLM Querying, while the make and model recognition pipeline has VLM Querying and an optional Self-Reflection Module.


\subsection{License Plate Recogntion}
\begin{figure}[!t]
  \centering
  \includegraphics[width=\textwidth]{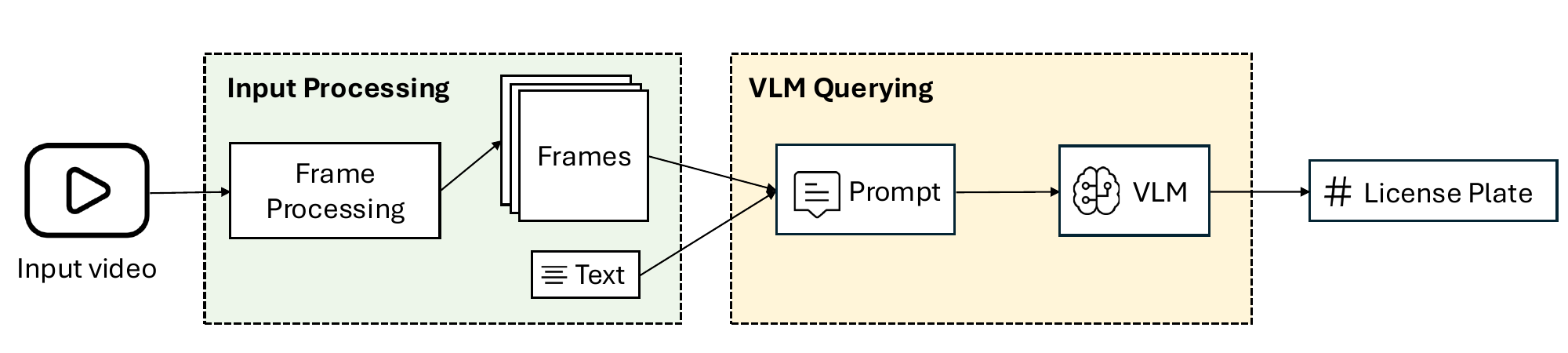}
  \caption{\textbf{Overview of the Vision–Language Pipeline for License Plate Recognition}} 
  \label{fig:framework}
\end{figure}

Our license plate recognition framework processes an input video to extract license plate information through two main stages: (1) Input Processing and (2) VLM querying, as illustrated in Figure \ref{fig:framework}. The Input Processing stage (green-shaded box) involves selecting the most informative and high-quality frames from the video to enhance recognition accuracy while significantly reducing the input size to the VLM. In the VLM Querying stage (yellow-shaded box), these selected frames are paired with carefully engineered textual prompts to form a unified multimodal input. This input is then passed to a VLM, which interprets both the visual and textual information to generate license plate.

\subsubsection{Input Processing}
\label{sec:input_processing}
The goal of the Input Processing stage is to select frames from the input video that are sharp, well-exposed, informative, and suitable for extracting license plate information. For example, if a license plate appears in multiple frames, it suffices to use only one frame that best captures the full license plate. This reduces the input size to the VLM by orders of magnitude compared to using all video frames. Moreover, selecting high-quality frames improves extraction accuracy since poor-quality images may lead to errors, such as misreading license plates or predicting incorrect vehicle types.

To identify the most informative frames from a video, we use two perceptual image quality metrics: CLIP-IQA (Contrastive Language-Image Pre-Training Image Quality Assessment) \cite{wang2022exploringclipassessinglook} and BRISQUE (Blind/Referenceless Image Spatial Quality Evaluator) \cite{6272356}. Both scoring methods rank frames based on perceived quality, allowing us to prioritize frames likely to yield accurate recognition results.

\begin{itemize}[leftmargin=2em]
    \item 
    \textbf{CLIP-IQA} is a no-reference image quality assessment metric that leverages CLIP embeddings. It computes similarity between an image’s CLIP embedding and a reference quality embedding (e.g., ``a high-quality photo'' or ``a blurry image'') as:
\begin{equation}
\text{CLIP-IQA}(x) = \cos\left( f_{\text{CLIP}}(x), f_{\text{CLIP}}(t_{\text{ref}}) \right)
\end{equation}
where \( f_{\text{CLIP}}(\cdot) \) denotes the CLIP embedding function and \( t_{\text{ref}} \) is a textual prompt representing good or poor quality. This score captures perceptual similarity and correlates well with human judgments of image quality \cite{wang2022exploringclipassessinglook}.
\item
\textbf{BRISQUE} is a handcrafted feature-based method that uses natural scene statistics (NSS) in the spatial domain to predict perceived quality. It models image distortions by extracting statistical features from locally normalized luminance coefficients and fits them into a support vector regressor trained on human opinion scores. BRISQUE is effective in assessing common distortions, such as noise, blur, and compression artifacts without requiring a reference image \cite{6272356}.
\end{itemize}

\subsubsection{Vision-Language Model Querying}
A multimodal prompt that consists of a composite image and a text prompt (see below) is used to extract a license plate from a video. For every sample, the same multimodal prompt will be used. The license plates that were previously extracted in the Input Processing Module are combined to create the composite image. A VLM receives this multimodal input and interprets the textual instructions and visual content together.

Additionally, three prompting strategies are compared when querying the VLMS. In the first strategy, a single query is issued, and the model returns its best prediction. In the second strategy, a single query is issued, and the model provides its top three predictions in one response. In the third strategy, the model is queried three separate times using the same prompt, and the prediction is considered correct if any of the three responses matches the ground truth. These strategies are evaluated to examine the impact of prompt design on the reliability and accuracy of model predictions.

\subsubsection{Text Prompt for License Plate Recognition}
\begin{tcolorbox}
\label{prompt:license_plate}
You are given multiple images of the same license plate. Each row shows one image of the license plate. Identify the license plate and predict at least six characters.

Do not include special characters such as colons or dashes.

The EasyOCR output is \texttt{\{easy\_ocr\}}. You may use this information to assist your prediction.

Return only a JSON object with the key \texttt{license\_plate} and the value set to the predicted license plate.
\end{tcolorbox}

\subsection{Make and Model Recognition}

\subsubsection{VLM Querying}
\begin{figure}[!t]
  \centering
  \includegraphics[width=\textwidth]{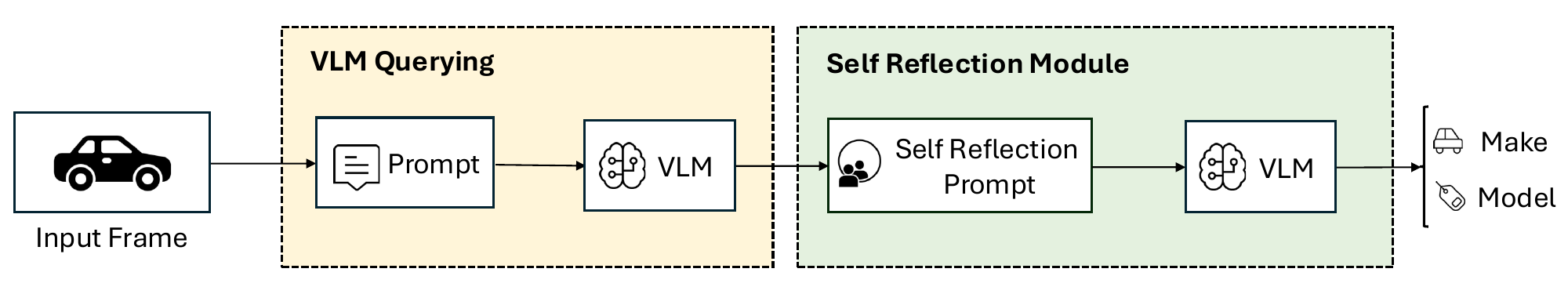}
  \caption{\textbf{An Overview of the Vision–Language Pipeline for Make and Model Recognition}} 
  \label{fig:framework_mmr}
\end{figure}

Figure~\ref{fig:framework_mmr} presents the framework proposed for make and model recognition. The vision-language model querying component follows the same procedure as in the license plate recognition framework, with the only difference being the use of a modified input prompt (see \emph{Initial Prediction Prompt}). In this case, frame selection based on image quality is not applied because make and model recognition is generally more robust to image blur and partial occlusions. An optional self-reflection module is included in the framework, and its functionality is described in greater detail in the subsequent section.

\subsubsection{Initial Prediction Prompt}

\begin{tcolorbox}
\label{prompt:initial}
Based on the given image, determine the make and model of the car from the following options: \{\textit{car\_options}\}. Output ONLY a JSON object with keys \texttt{make} and \texttt{model}.
\end{tcolorbox}

\subsubsection{Self-Reflection Module}

\begin{figure}[!t]
  \centering
  \includegraphics[width=\textwidth]{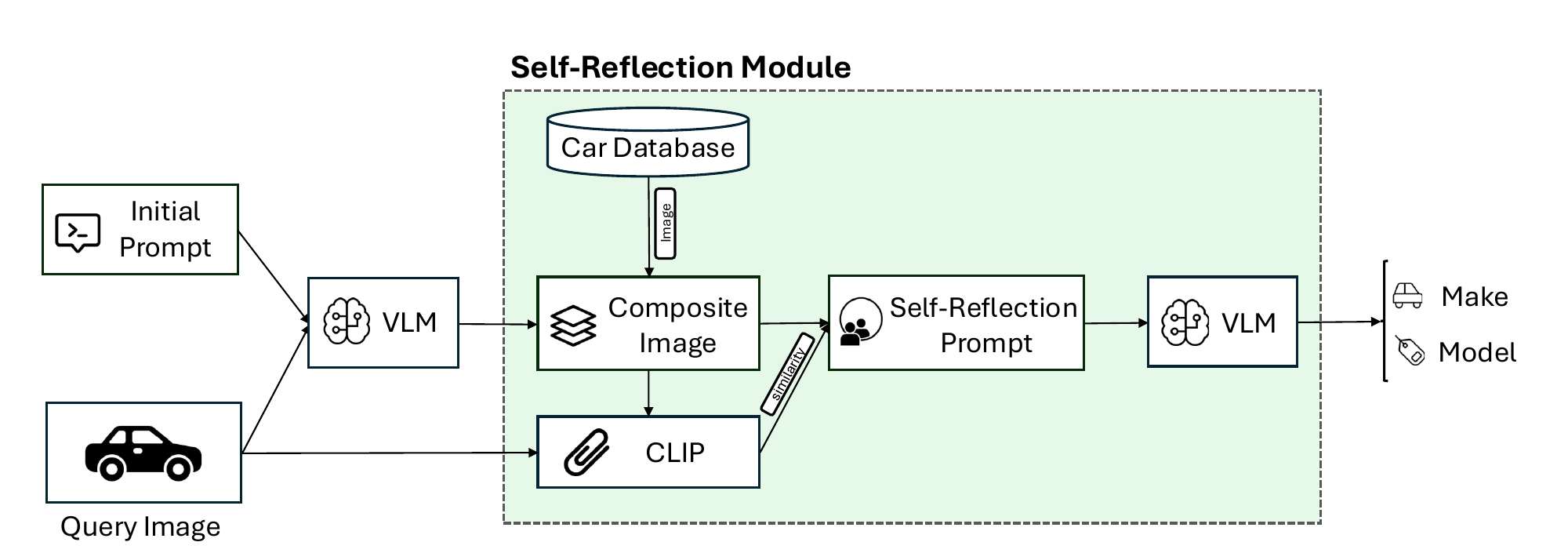}
  \caption{\textbf{Detailed Illustration of the self-reflection Module}}
  \label{fig:self_reflection}
\end{figure}

Although VLMs can produce structured predictions when given high-quality frames and well-crafted prompts, their initial outputs may still contain errors. This is especially likely when dealing with vehicles that have visually similar features or when the input image lacks distinctive details. To enhance prediction reliability, we introduce a self-reflection module (Figure~\ref{fig:self_reflection}). This additional stage enables the model to reconsider its initial output by comparing the query image to a visually similar reference image retrieved from a curated dataset. In this way, the self-reflection module builds directly on the earlier stages and improves the overall robustness of the pipeline.
To support the self-reflection module, we curated a reference dataset that allows the VLM to compare its initial prediction against a visually similar ground-truth image.  
To construct the dataset, we initiated an image crawling process from the web. As a preprocessing step, we removed the image backgrounds using the Segment Anything Model (SAM) \cite{kirillov2023segment}. Since SAM is class-agnostic and does not take semantic categories as input, we assumed that the central region of each image contained at least one pixel belonging to the vehicle. We curated a collection of 134 distinct make and model classes. For each image, we applied background removal and converted it to a black-and-white format. Additionally, vehicles were cropped and resized to ensure a consistent scale across all samples, facilitating uniform visual comparison.
\label{sec:dataset_collection}


Given a query image and an initial prompt, we first query a VLM to predict the make and model of the vehicle from a predefined set of options. Based on the initial prediction, we retrieve the corresponding vehicle image from the reference dataset described earlier. We then compute the visual similarity between the query image and the retrieved image. The visual similarity measure is computed using CLIP~\cite{radford2021learningtransferablevisualmodels}. Each image is processed by CLIP's image encoder to produce a high-dimensional embedding vector representing its visual characteristics. The cosine similarity between the embedding of the query image and that of the retrieved image from the reference dataset is then calculated, yielding a similarity score in the range \([0,1]\).

Self-reflection prompt comprises a textual prompt and a composite image. The textual prompt includes the similarity score between the query and retrieved images, the model's initial prediction, a predefined similarity threshold, and the complete set of candidate vehicle makes and models. The composite image is generated by placing the query image on the left and the retrieved reference image on the right, separated by a red bar, as illustrated in Table~\ref{tab:self_reflection_predictions}. The VLM is then prompted again with this information and explicitly instructed to compare its initial prediction with the retrieved reference image. If the visual evidence indicates a discrepancy, the model is encouraged to revise its prediction. This two-step procedure enables the VLM to refine its output by leveraging additional visual context.

\begin{tcolorbox}[enhanced,breakable]
\label{prompt:self_reflection}
\raggedright

You are shown TWO images merged into one separated by a red bar:

• Left: the query vehicle we must identify.

• Right: a rear-view photo of \textit{\{guess\}}

\vspace*{5px}
\includegraphics[width=0.3\textwidth]
{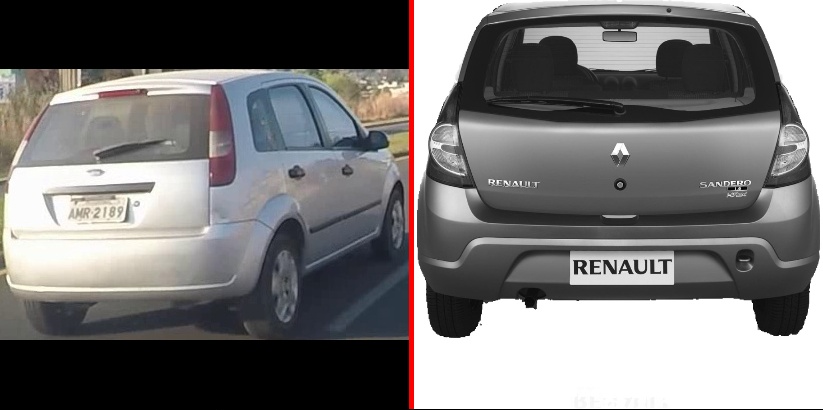}
\label{fig:plates}

\medskip

\raggedright
Your previous answer \textit{\{guess\}} received a similarity score of \textit{\{score\}} (target $\geq$ \textit{\{threshold\}}). Focus on shape, grille, taillights, and other cues. If they do not match, propose a different make/model than \textit{\{guess\}}. Choose one of these options: \{\textit{car\_options}\}

Output ONLY the JSON object with keys \texttt{make} and \texttt{model}, no explanations.
\end{tcolorbox}

\section{Experimental Setup}
We implemented our framework on an NVIDIA RTX 6000 Ada GPU using the YOLO model from \citet{Li2025smartphone}, which is fine-tuned for two epochs on smartphone videos for both vehicle and license plate detection tasks.
\subsection{Data}
We assessed both license plate and make and model recogntion using the UFPR-ALPR dataset~\cite{laroca2018robust}, which comprises 4,522 images of 150 distinct vehicles recorded in Brazil under dynamic capture conditions. The data were collected at 30 frames per second using consumer-grade cameras (GoPro Hero4 Silver, Huawei P9 Lite, and iPhone 7 Plus) mounted on a moving vehicle, while the target vehicles were also in motion. 
We also employed the dataset introduced by~\citet{Li2025smartphone}, comprising 24 smartphone videos recorded at 60 frames per second using an iPhone 12 from a roadside vantage point, resulting in approximately 12,300 frames. Each video is annotated with ground truth license plate strings and corresponding vehicle make and model labels, speed, and color, with annotations independently verified by two raters to ensure reliability. The dataset captures traffic on the campus of the University of Texas at Austin (101–153 E Dean Keeton St, Austin, TX 78712), where vehicles exhibit an average speed of 29.75 mph.
As the proposed framework operates entirely in a zero-shot setting, no dataset-specific training is required. The evaluation is conducted exclusively on the testing partitions of the considered datasets.

\subsection{Baselines}

For license plate recognition, we explored 24 different configurations by combining methods from three key components of our pipeline: two frame selection techniques (CLIP-IQA and BRISQUE), four VLMs (GPT-4o \cite{openai2024gpt4o}, Llama3.2-vision \cite{meta2024llama32}, LLaVA \cite{liu2023visual}, and MiniCPM-V \cite{yao2024minicpmv}), and three distinct prompting strategies (single call, three options, and three calls).

For make and model recognition, we evaluated the same set of VLMs used for license plate recognition on the smartphone dataset, then focused on GPT-4o and Llama 3.2-Vision to assess the impact of self-reflection. Results are reported for both the smartphone and UFPR-ALPR datasets, comparing performance with and without the self-reflection module.

\subsubsection{Frame Selection}
To identify the most informative frames from each video, we used two no-reference image quality assessment metrics: CLIP-IQA and BRISQUE. These methods score each frame based on perceived visual quality, allowing us to prioritize those likely to improve recognition accuracy.



\subsubsection{Vision-Language Model}
we evaluated four recent and widely available VLMs: GPT-4o, Llama-3.2-Vision, LLaVA, and MiniCPM-V. These models were selected due to their strong performance on multimodal benchmarks and accessibility through open-source release or API endpoints.

\subsubsection{Prompting Strategies}
Three prompting strategies are evaluated when querying VLMs for license plate recognition
%
%
\begin{itemize}
    \item \textbf{Single Call}: The VLM is queried once to provide its top prediction.
    \item \textbf{Three Options}: The VLM is queried once and instructed to return its three most probable predictions.
    \item \textbf{Three Calls}: The VLM is queried three separate times, each time requested to provide its single best prediction.
\end{itemize}

As a non-VLM baseline, we replicated the pipeline proposed by~\citet{Li2025smartphone}, which uses a fine-tuned YOLO detector followed by EasyOCR for license plate recognition, with post-processing to select the most frequent prediction.

\subsubsection{self-reflection Module}
We conducted vehicle make and model recognition both with the self-reflection module described earlier and without it in order to evaluate its effect on overall performance.

\section{Results}

\begin{figure*}[!t]
    \centering
    \begin{subfigure}{0.48\textwidth}
        \centering
        \includegraphics[width=\linewidth]{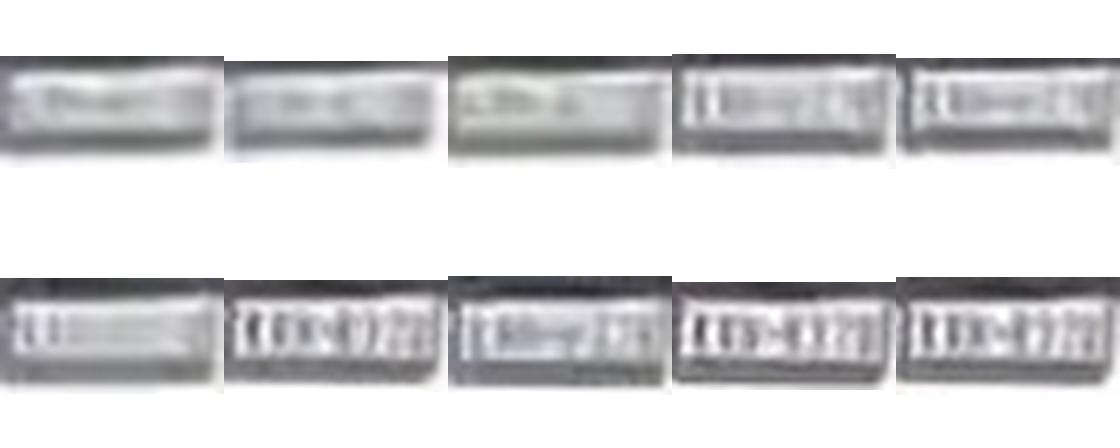}
        \caption{BRISQUE — lowest 10 scores}
        \label{fig:brisque_low}
    \end{subfigure}
    \hfill
    \begin{subfigure}{0.48\textwidth}
        \centering
        \includegraphics[width=\linewidth]{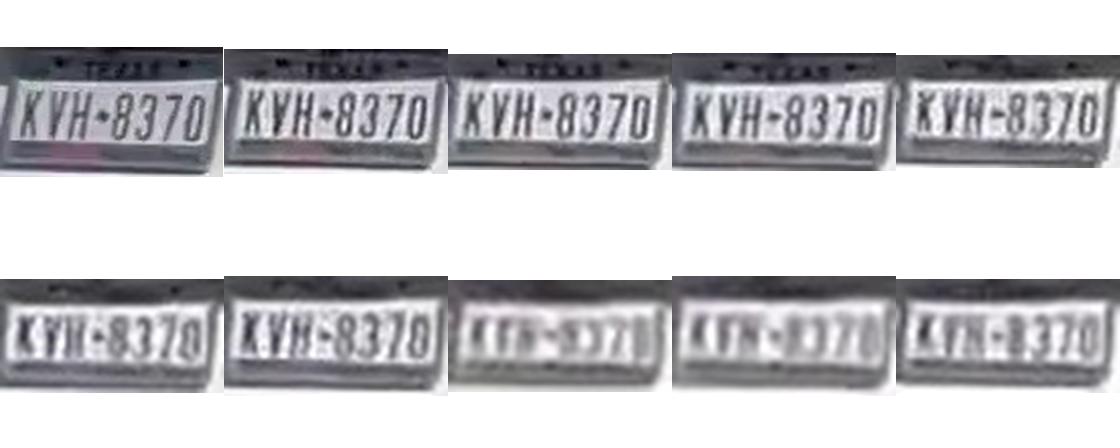}
        \caption{BRISQUE — highest 10 scores}
        \label{fig:brisque_high}
    \end{subfigure}

    \vspace{0.6em} 

    \begin{subfigure}{0.48\textwidth}
        \centering
        \includegraphics[width=\linewidth]{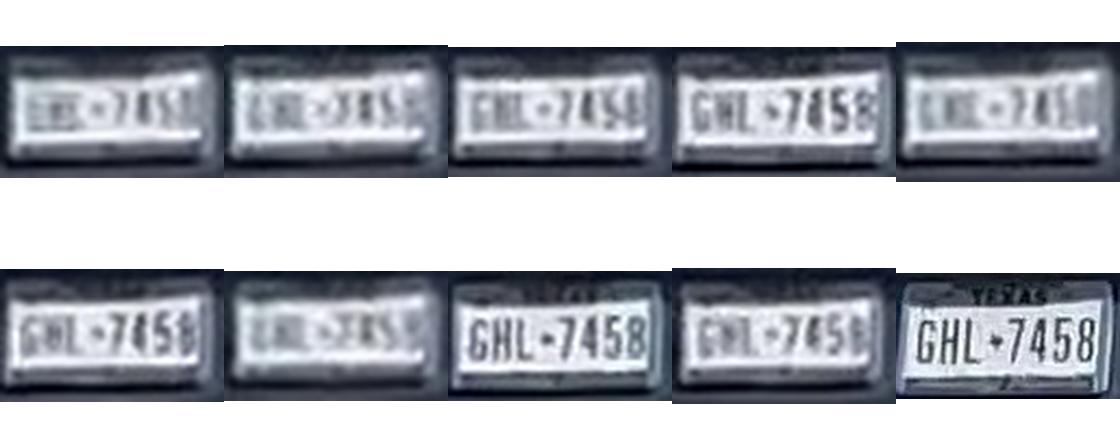}
        \caption{CLIP‑IQA — lowest 10 scores}
        \label{fig:clip_low}
    \end{subfigure}
    \hfill
    \begin{subfigure}{0.48\textwidth}
        \centering
        \includegraphics[width=\linewidth]{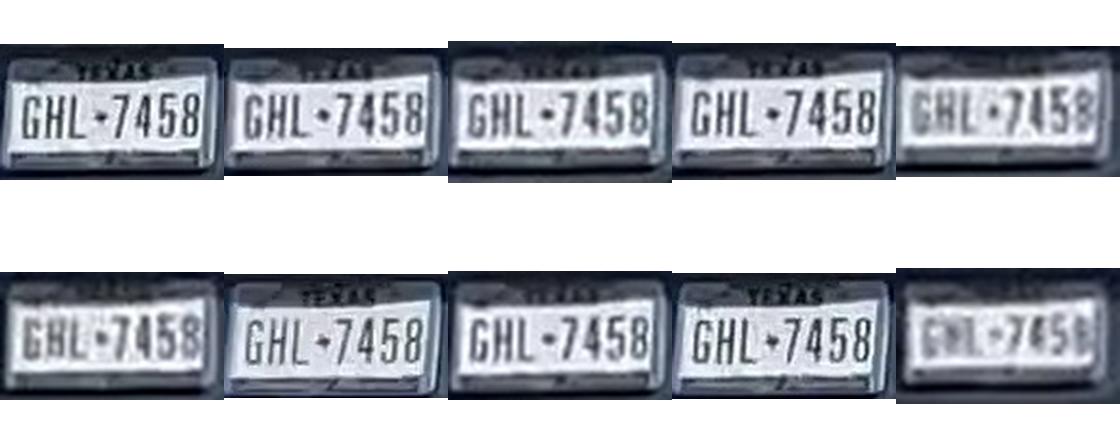}
        \caption{CLIP‑IQA — highest 10 scores}
        \label{fig:clip_high}
    \end{subfigure}

    \caption{\textbf{Quality Extremes for Two Image‑Quality Assessors}}
    \label{fig:iqa_comparison}
\end{figure*}

Figure~\ref{fig:iqa_comparison} illustrates the quality extremes identified by two image quality assessment methods. The top row corresponds to frames evaluated using BRISQUE, while the bottom row corresponds to frames evaluated using CLIP-IQA. Each panel presents the ten frames with the lowest or highest quality scores for the same vehicle, arranged in a grid with two rows and five columns.

Table~\ref{tab:vlm_on_lpr_prediction_comparison_alpr} reports zero‑shot license–plate recognition accuracy on the UFPR‑ALPR benchmark as a function of the frame‑quality metric, VLM, and the prompting strategy.

Independent querying is consistently beneficial, but its impact varies by model capacity.  
For the strongest model, GPT‑4o, accuracy rises from 83.1\% with a single call to 86.4\% with either three options or three calls (using CLIP‑IQA), a modest 3‑point gain that effectively saturates performance.  
Mid‑tier MiniCPM‑V benefits markedly, CLIP‑IQA accuracy nearly doubles, climbing from 28.8\% (single call) to 54.2\% (three calls).  
The weakest model, Llava, exhibits only limited recovery, never surpassing 22\%, and even collapses to 0\% under the three‑option prompt, suggesting susceptibility to prompt‑format changes when the underlying visual grounding is poor.

CLIP‑IQA generally produces the highest scores for GPT‑4o and MiniCPM‑V, whereas BRISQUE matches or slightly exceeds CLIP‑IQA for Llama 3.2‑Vision under the three‑option strategy (84.8\% for both metrics).  
No single metric dominates across models, indicating that either perceptual proxy can serve as an effective pre‑filter when paired with a capable VLM.

\begin{table}[!b]
    \caption{\textbf{License Plate Recognition Accuracy On UFPR-ALPR Dataset}}
    \label{tab:vlm_on_lpr_prediction_comparison_alpr}
    \begin{center}
        \begin{tabular}{llccc}
            \hline
            Metric & Model & Single Call & Three Options & Three Calls \\
            \hline
            \multirow{4}{*}{CLIP-IQA} 
                & GPT-4o & 83.05\% (49/59) & 86.44\% (51/59) & 86.44\% (51/59) \\
                & Llama3.2-vision & 66.10\% (39/59) & 84.75\% (50/59) & 76.27\% (45/59) \\
                & Llava & 16.95\% (10/59) & 00.00\% (0/59) & 20.34\% (12/59) \\
                & MiniCPM-V & 28.81\% (17/59) & 50.85\% (30/59) & 54.24\% (32/59) \\
            \hline
            \multirow{4}{*}{BRISQUE} 
                & GPT-4o & 77.97\% (46/59) & 84.75\% (50/59) & 84.75\% (50/59) \\
                & Llama3.2-vision & 62.71\% (37/59) & 84.75\% (50/59) & 69.49\% (41/59) \\
                & Llava & 22.03\% (13/59) & 00.00\% (0/59) & 22.03\% (13/59) \\
                & MiniCPM-V & 42.37\% (25/59) & 44.07\% (26/59) & 50.85\% (30/59) \\
            \hline
            \multicolumn{5}{l}{\textit{Baseline (Non-VLM):} 46\% accuracy reported by \citet{Li2025smartphone}} \\
            \hline
        \end{tabular}
    \end{center}
\end{table}
\begin{table}[!b]
    \caption{\textbf{License Plate Recognition Accuracy On Smartphone Dataset}}
    \label{tab:vlm_on_lpr_prediction_comparison}
    \begin{center}
        \begin{tabular}{llccc}
            \hline
            Metric & Model & Single Call & Three Options & Three Calls \\
            \hline
            \multirow{4}{*}{CLIP-IQA} 
                & GPT-4o & 83.33\% (20/24) & 87.50\% (21/24) & 91.67\% (22/24) \\
                & Llama3.2-vision & 91.67\% (22/24) & 91.67\% (22/24) & 87.50\% (21/24) \\
                & Llava & 25.00\% (6/24) & 25.00\% (6/24) & 29.17\% (7/24) \\
                & MiniCPM-V & 54.17\% (13/24) & 54.17\% (13/24) & 70.83\% (17/24) \\
            \hline
            \multirow{4}{*}{BRISQUE} 
                & GPT-4o & 79.17\% (19/24) & 87.50\% (21/24) & 87.50\% (19/24) \\
                & Llama3.2-vision & 87.50\% (21/24) & 87.50\% (21/24) & 91.67\% (22/24) \\
                & Llava & 33.33\% (8/24) & 33.33\% (8/24) & 33.33\% (8/24) \\
                & MiniCPM-V & 58.33\% (14/24) & 54.17\% (13/24) & 79.17\% (19/24) \\
            \hline
            \multicolumn{5}{l}{\textit{Baseline (Non-VLM):} 29.7\% top-10 accuracy, reported by \citet{Li2025smartphone}} \\
            \hline
        \end{tabular}
    \end{center}
\end{table}

Table \ref{tab:vlm_on_lpr_prediction_comparison} contrasts the influence of the frame‑quality metric, the VLM, and the prompting strategy on smartphone data. For the two strongest VLMs, i.e., GPT‑4o and Llama 3.2‑Vision, zero‑shot inference already delivers competitive results (83.3\% and 91.7\%, respectively, when CLIP‑IQA is used), but issuing three independent calls pushes accuracy to 91.7\% and 91.7\%, reflecting absolute gains of 8–13 percentage points. The mid-tier MiniCPM-V model exhibits notable performance gains with repeated querying, improving from approximately \(54\%\) and \(58\%\) under a single call to \(71\%\) and \(79\%\) with three calls, respectively. In contrast, the lowest-performing Llava model shows minimal sensitivity to additional queries, with accuracy remaining in the range of roughly \(25\text{--}33\%\). Comparing the two frame‑quality metrics, CLIP‑IQA yields the best score for GPT‑4o, whereas BRISQUE leads for Llama 3.2‑Vision; thus, no single metric dominates, but either is sufficient to drive high performance when paired with a strong VLM.




\begin{figure}[!t]
  \centering

  \begin{subfigure}[t]{\textwidth}
    \centering
    \includegraphics[width=0.8\linewidth]{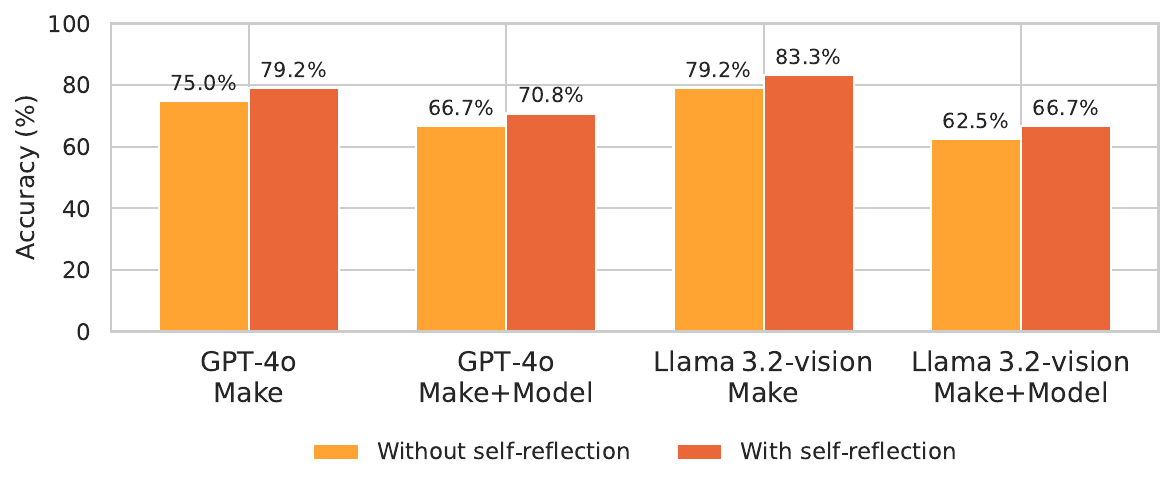}
    \caption{Smartphone dataset}
    \label{fig:accuracy_self_reflection:smartphone}
  \end{subfigure}

  \vspace{0.75ex} 

  \begin{subfigure}[t]{\textwidth}
    \centering
    \includegraphics[width=0.8\linewidth]{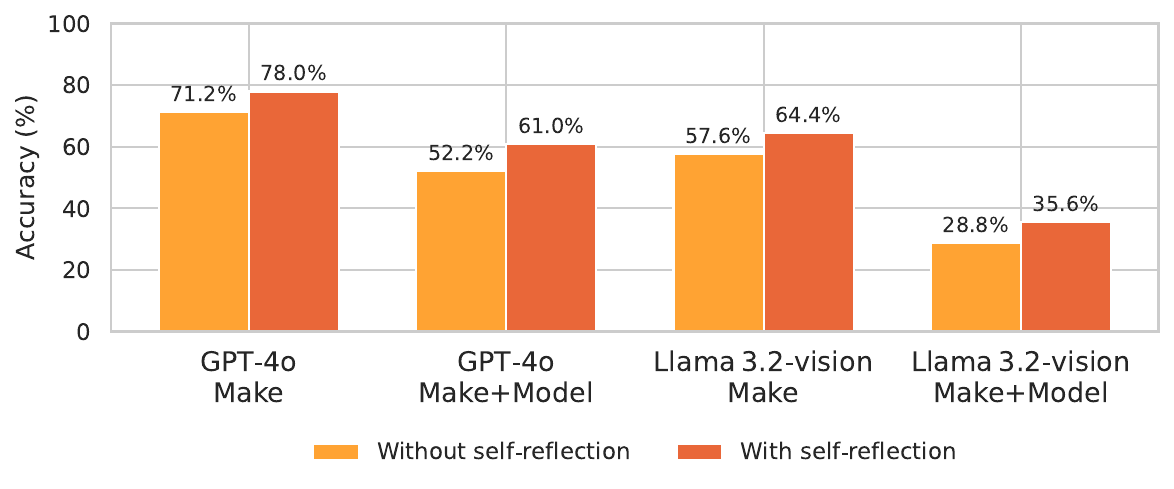}
    \caption{UFPR-ALPR dataset}
    \label{fig:accuracy_self_reflection:ufpr}
  \end{subfigure}

  \caption{\textbf{Accuracy Comparison With and Without Self-Reflection Module} (The baseline accuracy is 48.60\% for make and 16.89\% for make + model reported by \citep{Li2025smartphone})}
  \label{fig:accuracy_self_reflection}
\end{figure}

When the task requires simultaneous make and model recognition, accuracy drops for every model, reflecting the added fine‑grained visual complexity.  
Using three-call prompting helps recover much of the lost performance. GPT-4o and Llama~3.2-Vision both reach \(70.8\%\), improving by 8 and 4 percentage points compared to their single-call results. MiniCPM-V also increases from \(20.8\%\) to \(29.2\%\), whereas Llava shows only a slight improvement from \(16.7\%\) to \(20.8\%\).  
Importantly, the best VLM configurations surpass the non‑VLM baseline by more than four‑fold (70.8\%$\rightarrow$16.9\%), demonstrating that, even without fine‑tuning, contemporary VLMs can deliver state‑of‑the‑art performance on challenging recognition tasks.

Representative examples of both successful and failed predictions are provided in Table~\ref{tab:veh_pred_vs_gt}, illustrating the strengths and limitations of the proposed framework under varying image conditions.

Figure~\ref{fig:accuracy_self_reflection} summarizes the effect of the self-reflection module on make and make+model recognition accuracy. Each pair of bars shows the improvement (or degradation) in performance for a specific model and dataset after applying self-reflection. 

Figure~\ref{fig:accuracy_self_reflection} (a) shows the results for the smartphone dataset. The self-reflection step improves accuracy consistently across all evaluated VLMs, with gains ranging from approximately four to five percentage points for both make-only and make+model classification tasks. This pattern indicates that the mechanism provides a uniform benefit, regardless of the underlying model architecture.

Figure~\ref{fig:accuracy_self_reflection} (b) presents the results for the UFPR-ALPR dataset. A similar trend is observed; each model achieves slightly higher accuracy after the self-reflection step, and the magnitude of improvement is relatively stable across models. These results suggest that self-reflection systematically enhances the reliability of predictions on both datasets, although the improvements remain modest in size.

Table~\ref{tab:self_reflection_predictions} illustrates individual examples in which the self-reflection process adjusts an initial misclassification. These cases highlight how comparing the query image against retrieved reference images can guide the model toward more accurate predictions.

\begin{table*}[t]
  \centering
  \caption{\textbf{Vehicle Prediction–Ground Truth Comparison}}
  \label{tab:veh_pred_vs_gt}
  \resizebox{\textwidth}{!}{
  \begin{tabular}{|
      C{0.32\textwidth}|
      C{0.22\textwidth}|
      C{0.46\textwidth}|}
      \hline
      \textbf{Vehicle} & \textbf{Top-3 plates} & \textbf{Prediction vs Ground Truth} \\
      \hline
      \includegraphics[width=\linewidth]{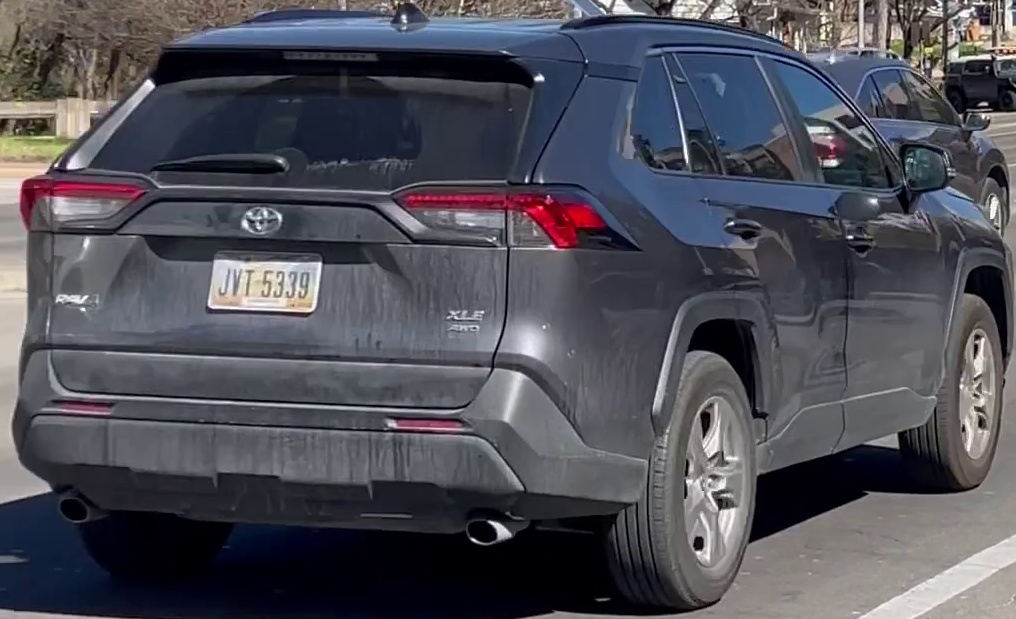} &
      \includegraphics[width=0.5\linewidth]{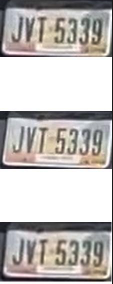} &
      \shortstack{JVT5339\,Toyota\,RAV4\\(same)} \\
      \hline
      \includegraphics[width=\linewidth]{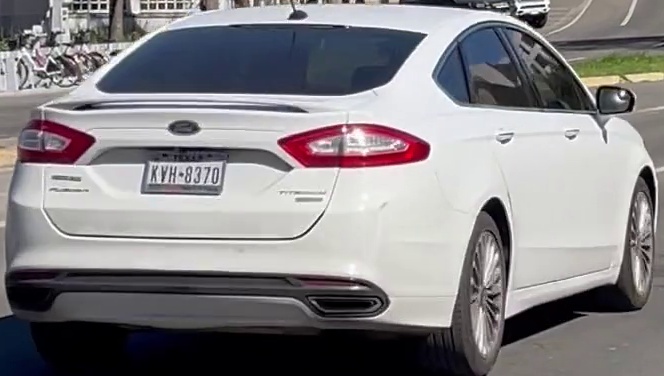} &
      \includegraphics[width=0.5\linewidth]{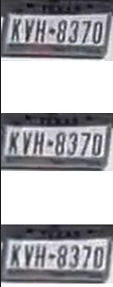} &
      \shortstack{KVH8370\,Ford\,Fusion\\(same)} \\
      \hline
      \includegraphics[width=\linewidth]{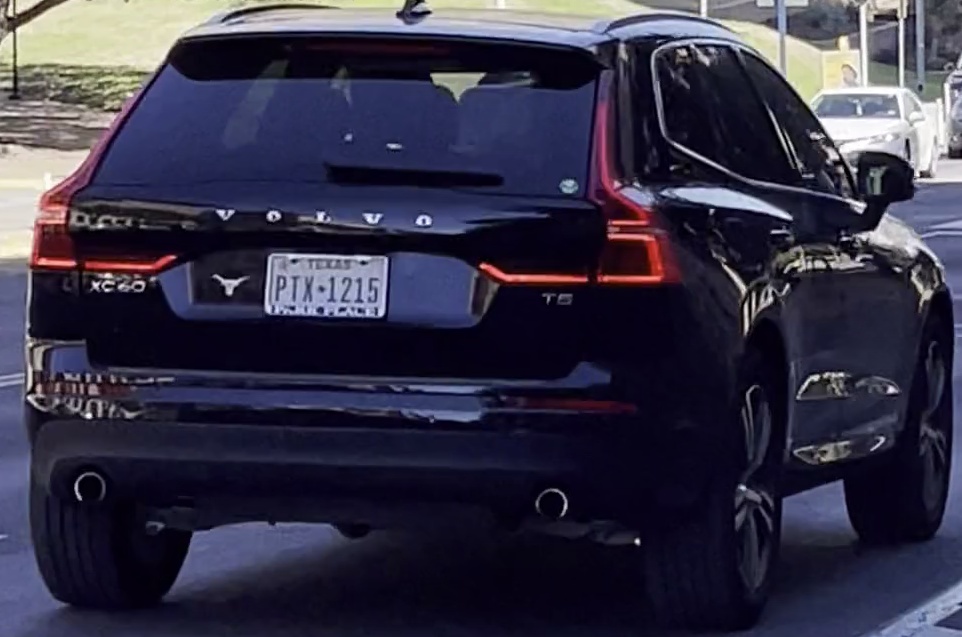} &
      \includegraphics[width=0.5\linewidth]{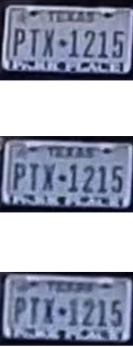} &
      \shortstack{PTX1215\,Volvo\,XC\textcolor{red}{90}\\vs\\PTX1215\,Volvo\,XC\textcolor{red}{60}} \\
      \hline
      \includegraphics[width=\linewidth]{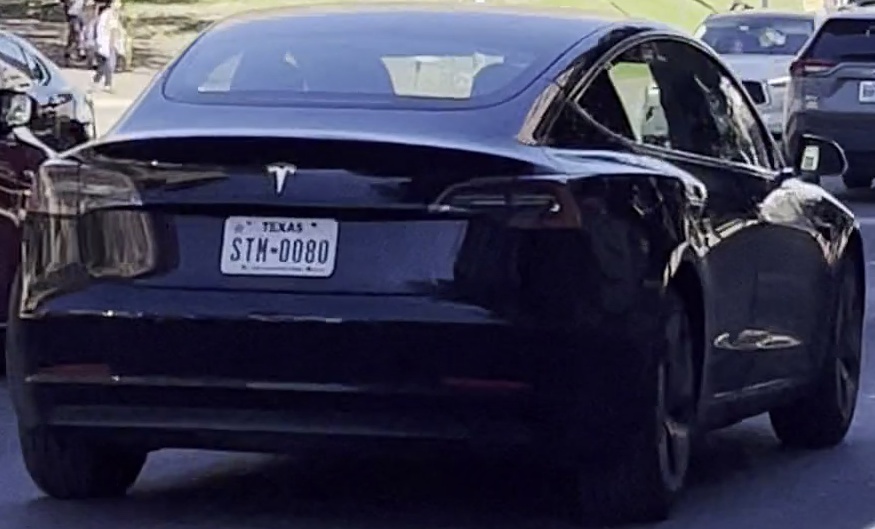} &
      \includegraphics[width=0.5\linewidth]{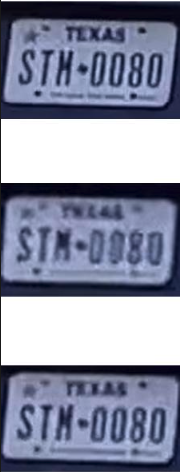} &
      \shortstack{ST\textcolor{red}{H}0080\,Tesla\,Model\,3\\vs\\ST\textcolor{red}{M}0080\,Tesla\,Model\,3} \\
      \hline
  \end{tabular}}
\end{table*}

\begin{table*}[t]
    \caption{\textbf{Self-Reflection Prediction Comparison}}
    \centering
    \resizebox{\textwidth}{!}{
    \begin{tabular}{|
        C{0.60\textwidth}|  
        C{0.20\textwidth}|  
        C{0.20\textwidth}|} 
        \hline
        \textbf{Composite image} &
        \textbf{Step 1 prediction} &
        \textbf{Step 2 prediction} \\
        \hline
        \includegraphics[width=\linewidth]{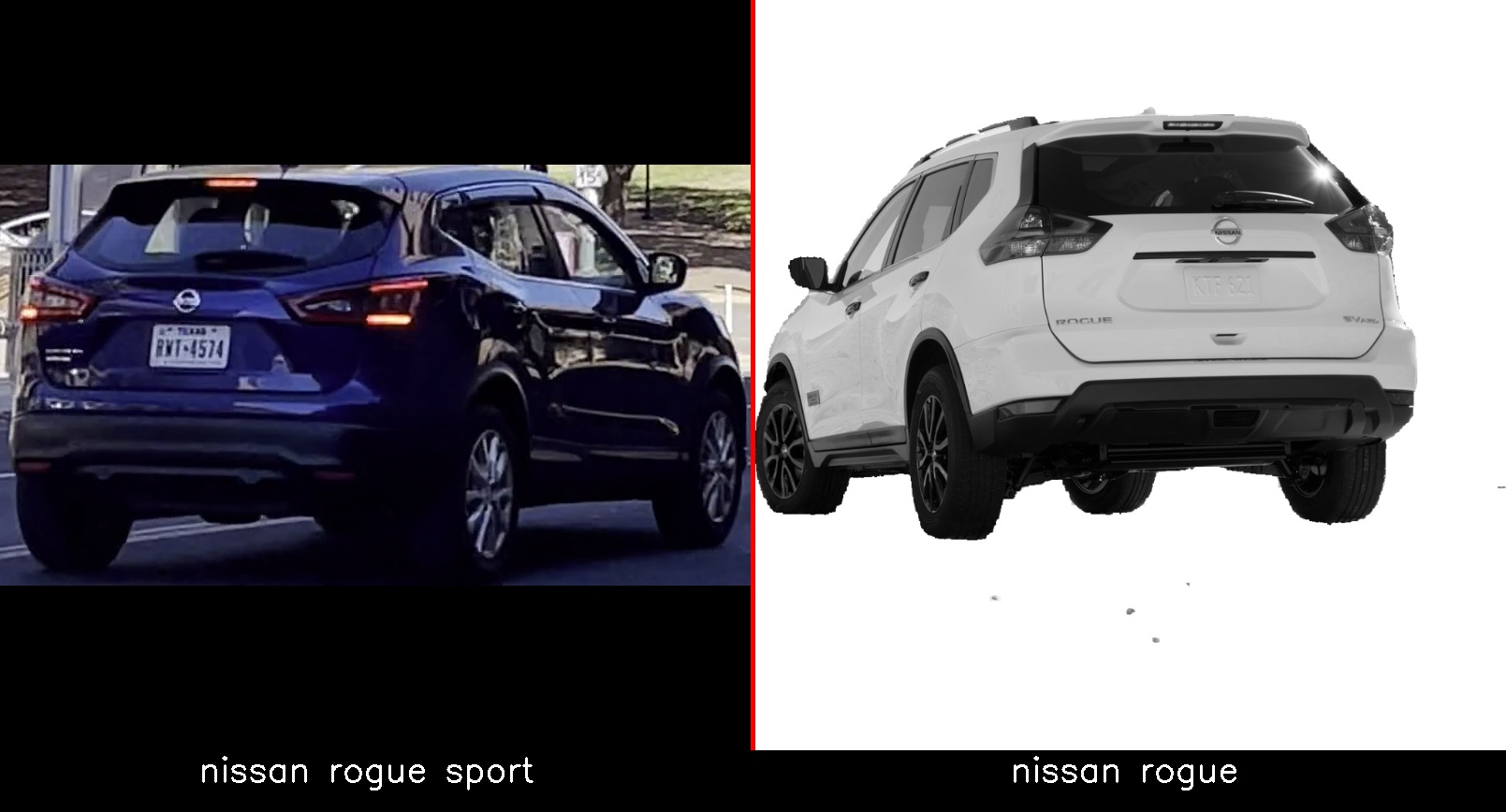} &
        \shortstack{Nissan\\ Rogue} &
        \shortstack{Nissan\\ Rogue Sport} \\
        \hline
        \includegraphics[width=\linewidth]{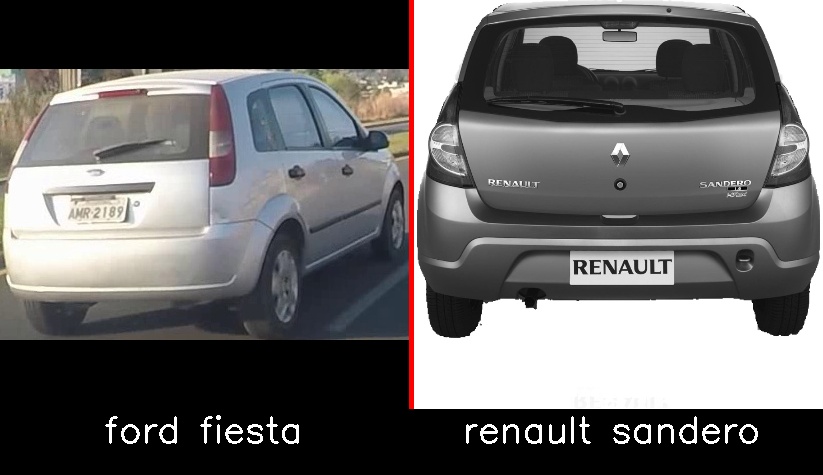} &
        \shortstack{Renault\\ Sandero} &
        \shortstack{Ford\\ Fiesta} \\
        \hline
        \includegraphics[width=\linewidth]{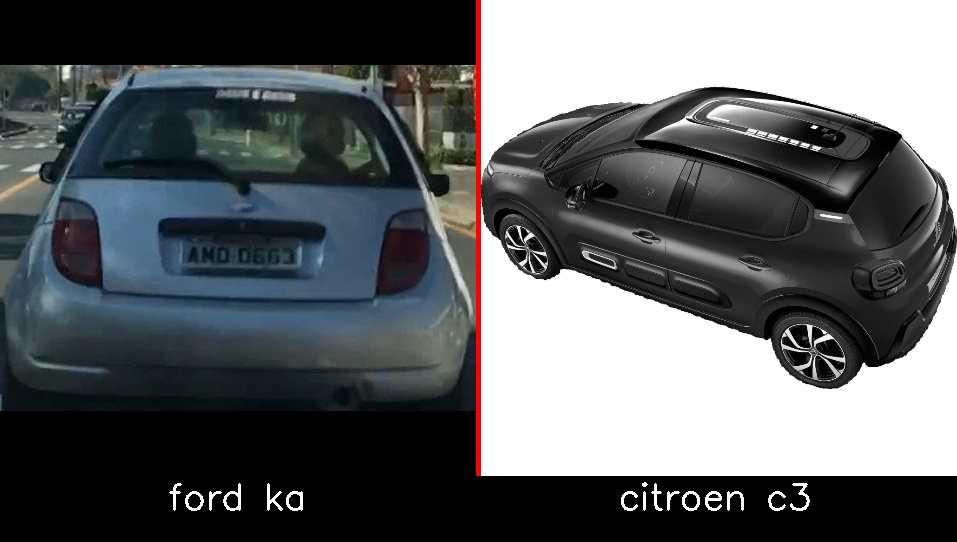} &
        \shortstack{Citroen\\ C3} &
        \shortstack{Ford\\ Ka} \\
        \hline
    \end{tabular}
    }
    \label{tab:self_reflection_predictions}
\end{table*}

\section{Conclusion}
This study demonstrates the viability of using VLMs for robust, end-to-end license plate and vehicle attribute recognition in unconstrained video settings, such as those captured by smartphones. By leveraging a unified pipeline that combines object detection, frame selection through perceptual quality metrics, and prompt-based VLM querying, our method significantly outperforms traditional approaches, achieving up to 91.7\% plate recognition accuracy and 70.83\% make/model accuracy. Notably, our zero-shot approach eliminates the need for fine-tuning, making it more accessible for deployment on resource-limited devices. The results highlight the potential of open-source VLMs like Llama-3.2-Vision to match proprietary models in accuracy while enabling on-device processing at zero cost, offering a scalable and flexible solution for future intelligent transportation systems.

\subsection{Limitations}
Despite the strong zero-shot performance demonstrated in this study, several limitations remain that should be considered in future work. The pipeline relies heavily on the quality and representativeness of the selected frames, and extreme motion blur, severe occlusions, or highly unusual viewing angles can still undermine both object detection and VLM prompting, leading to misinterpretations or hallucinations. While off-the-shelf VLMs eliminate the need for task-specific training, they may still produce unpredictable errors, particularly when interpreting stylized or non-standard license plates, such as novelty plates or those containing foreign characters that are underrepresented in pretraining data. Furthermore, the end-to-end latency and computational requirements, although acceptable on high-end GPUs, may pose challenges for deployment on resource-constrained devices. Batching multiple calls to remote APIs can also introduce variable network delays and additional costs. Proprietary VLMs, such as GPT-4o \cite{openai2024gpt4o} achieved the highest accuracy in our experiments, yet their cost may become a significant barrier for large-scale deployments. The self-reflection module further increases the number of API calls, raising processing costs to approximately \$0.17 for the 24-sample smartphone dataset and \$0.42 for the 59-sample UFPR-ALPR dataset in our evaluation. These figures already reflect one of the most affordable APIs available, whereas more advanced alternatives , such as o1-pro can be up to sixty times more expensive. Finally, our evaluation is based on two datasets that, while diverse, cannot fully represent the global variability in license plate designs, lighting conditions, and camera hardware that real-world citizen enforcement scenarios may encounter.


\subsection{Future Directions}
To address these challenges, future work could explore several avenues. From a systems perspective, optimizing the pipeline for on-device execution (for example via model quantization, distillation, or pruning) would broaden applicability to smartphones and edge cameras. Expanding our benchmarks to include publicly available dashcam datasets and international plate collections will help quantify generalization. Tight integration with enforcement workflows (including automated database lookups, privacy-preserving logging, and human-in-the-loop verification) will also be essential to translate these technical advances into real-world safety and compliance gains.

\clearpage

\section{Declaration of Generative AI and AI-Assisted Technologies in the Writing Process}
During the preparation of this manuscript, the authors made use of ChatGPT as a tool for language assistance. After utilizing these tools, the authors thoroughly reviewed and revised the content as necessary and take full responsibility for the final version of the published article.

\section{Author Contribution Statement}

The authors confirm contribution to the paper as follows: 
\textbf{study conception and design:} Pouya Parsa, Keya Li, Kara M. Kockelman, Seongjin Choi; 
\textbf{data collection:} Pouya Parsa, Keya Li, Kara M. Kockelman; 
\textbf{analysis and interpretation of results:} Pouya Parsa, Seongjin Choi; 
\textbf{draft manuscript preparation:} Pouya Parsa, Seongjin Choi. 
All authors reviewed the results and approved the final version of the manuscript.

\bibliographystyle{trb}
\bibliography{trb_template}
\end{document}